%% file: ISEDA2023_Routability.tex
\documentclass[conference,9pt]{IEEEtran}
\IEEEoverridecommandlockouts
\input{./texlib/ieee_setting}

\newcommand\blfootnote[1]{%
  \begingroup
  \renewcommand\thefootnote{}\footnote{#1}%
  \addtocounter{footnote}{-1}%
  \endgroup
}
\graphicspath{{./figs/}}

\begin{document}

\makeatletter
 % 作者居中
\newcommand{\linebreakand}{%
  \end{@IEEEauthorhalign}
  \hfill\mbox{}\par
  \mbox{}\hfill\begin{@IEEEauthorhalign}
}
\makeatother                                                                                                     
% \acmISBN{978-1-4503-6725-7/19/06}

\title{
\textbf{
HybridNet: Dual-Branch Fusion of Geometrical and Topological Views for VLSI Congestion Prediction (Extended Abstract)
}}

\author{
  % \large 
  Yuxiang Zhao$^{1}$,
  Zhuomin Chai$^{2}$,  
  Yibo Lin$^{1, 3, 4*}$, 
  Runsheng Wang$^{1, 3, 4}$, 
  Ru Huang$^{1, 3, 4}$ \\
  $^1$School of Integrated Circuits, Peking University \\
  $^2$School of Physics and Technology, Wuhan University \\
  % $^3$Institute for Artificial Intelligence, Peking University \\
  $^3$Beijing Advanced Innovation Center for Integrated Circuits \\
  $^4$Institute of Electronic Design Automation, Peking University, Wuxi, China
  %{\normalsize \tt \{gzz,yibolin\}@pku.edu.cn}
}

\maketitle
\input{doc/abstract}
\blfootnote{$^*$Corresponding author: yibolin@pku.edu.cn}
\input{doc/intro}

\input{doc/algo}
\input{doc/result}
\input{doc/conclu}
\input{doc/acknowledgement}

\vspace{-.05in}
{
%\scriptsize
\small
\bibliographystyle{IEEEtran}
\bibliography{./ref/Top_sim,./ref/baseline}
% \bibliography{./ref/baseline}
}

\end{document}

%% file: texlib/ieee_setting.tex
\usepackage{algpseudocode}    % new algorithm package
\usepackage{algorithm}
\usepackage{amsmath}
\usepackage{amssymb}
\usepackage{amsfonts}
\usepackage{booktabs}
\usepackage{multirow}
\usepackage{subfigure}
\usepackage{graphicx}         % include pdf figures
\usepackage{color}
\usepackage{cite}             % more citations in one bracket
\usepackage{comment}          % use comment
\usepackage{soul}             % use highlight command \hl{}
\soulregister\cite7
\soulregister\ref7
\soulregister\pageref7
\usepackage{amsthm}
\usepackage{etoolbox}         % commands \newtoggle, \toggletrue, \iftoggle
\usepackage{url}
\usepackage{nth}              % nth command
\usepackage{bm}               % bm command

\makeatletter
\let\OldStatex\Statex
\renewcommand{\Statex}[1][3]{%
  \setlength\@tempdima{\algorithmicindent}%
  \OldStatex\hskip\dimexpr#1\@tempdima\relax
}
\makeatother

%% file: doc/abstract.tex
\begin{abstract}
Accurate early congestion prediction can prevent unpleasant surprises at the routing stage, playing a crucial character in assisting designers to iterate faster in VLSI design cycles.
In this paper, we introduce a novel strategy to fully incorporate topological and geometrical features of circuits by making several key designs in our network architecture.
To be more specific, we construct two individual graphs (geometry-graph, topology-graph) with distinct edge construction schemes according to their unique properties.
We then propose a dual-branch network with different encoder layers in each pathway and aggregate representations with a sophisticated fusion strategy.
Our network, named HybridNet, not only provides a simple yet effective way to capture the geometric interactions of cells, but also preserves the original topological relationships in the netlist.
Experimental results on the ISPD2015 benchmarks show that we achieve an improvement of 10.9\% compared to previous methods.
\end{abstract}

% \vspace{-2mm}

%% file: doc/intro.tex
\section{Introduction}
\label{sec:Introduction}

As design complexity increases, efficient and accurate prediction of routing congestion is critical  to assist placement to achieve routability in the design flow. 
%
%%%
To better achieve this target, machine learning has been used to help predict congestion.
Existing methods can be divided into two parts, vision-based and graph-based methods, in terms of the model's input characteristics.
Vision-based methods treat the grid-level feature directly as pixel channels of an image, turning the prediction problem into an image-to-image translation task~\cite{isola2017image, chai2022circuitnet, routenet, liu2021global}.
~\cite{high-definition} use the popular Pix2Pix methods to accomplish this task 
by converting the hand-crafted features input image to make final predictions.
However, the fundamental feature representing the topological relationship in netlist data is overlooked in the pixel conversion process.
Graph-based methods focus on capturing node relationships by transforming the raw netlist data into a graph structure with geometric information.~\cite{congestionnet, congestionnetnn}.
CircuitGNN~\cite{yangversatile} uses a heterogeneous graph to link the cell-net interconnection and mixes the topology and geometric features together through each aggregation layer. This method confuses the message-passing function to distinguish the finer difference between two different types of features and will affect the prediction performance.

\begin{figure}
    \centering
    % \vspace{5mm}
    \includegraphics[width=0.43\textwidth]{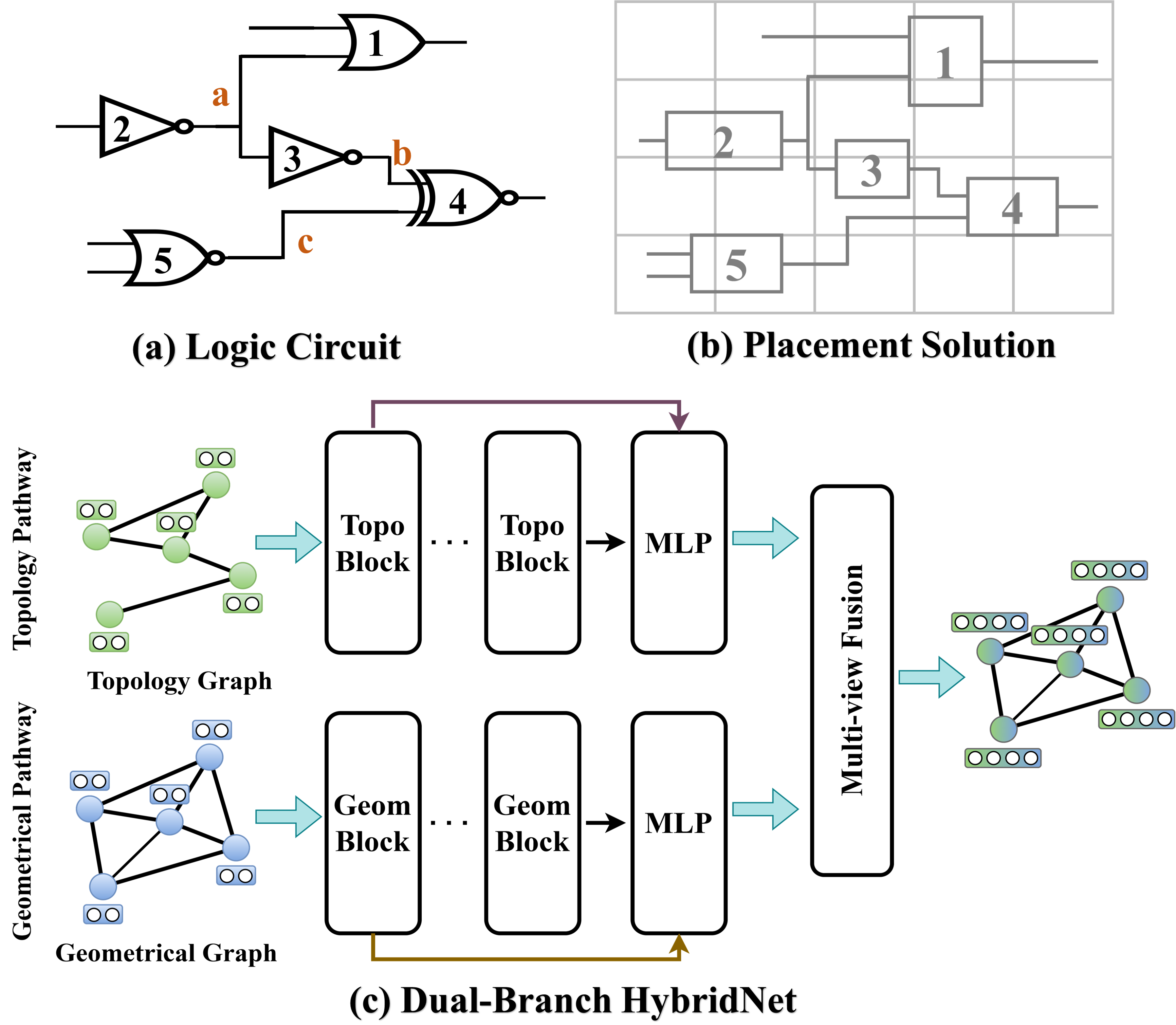}
    \caption{Circuit design with placement solution and the HybridNet architecture.}
    \label{fig:WirelengthOpRuntime}
    \vspace{-5mm}
\end{figure}

%%%
In this study, we aim to improve the representation of such netlist data for congestion prediction.
We're motivated by the similar characteristics shared by multiple individuals in the social network domain~\cite{survey}.
Individuals in real-world social networks are in contact with others through a variety of relationship types.
These relationships correspond to distinct views of the underlying network, which are naturally represented as multi-view graphs in each relationship type.
Such multi-view nature can also be found in VLSI circuits. 
Thus, we propose to consider the original netlist as a multi-view graph:
(i) The topology-graph is formed by the intrinsic configuration in the netlist logic structure, with each node containing features designed by human experts as described in the Methodology section.
(ii) The geometry-graph is built by leveraging the Delaunay triangulation algorithm for node interaction, where inter-node distances and coordinates are provided as extra features for edges and nodes respectively.
To this end, we introduce our framework, HybridNet, a dual-branch network consisting of two paths that operate separately on topological and geometric graphs.
The main contributions of our work are summarized as follows:

\begin{itemize}
    \item 
    We consider a circuit netlist as a multi-view graph and establish a topological-geometrical view graph structure for the inter-node relationship, which is exactly suited to represent the unique properties of the netlist compared to the single-view counterpart.
    \item 
    We propose HybridNet, a novel dual-branch network for learning multi-view graph representations that exploits a more promising netlist context through feature aggregation.
    \item 
    Experimental results on ISPD2015 benchmarks demonstrate the effectiveness of our method and achieve better accuracy 
    compared to previous works in the literature.
\end{itemize}

%% file: doc/algo.tex
\section{Methodology}
\label{sec:Algorithm}
\subsection{Multi-view Graph Construction}
As shown in Figure 1a, after the logic synthesis step we formalise a circuit design into a gate-level netlist. Cells and Nets are the two basic elements of the netlist, where Cells refer to the logic gate and Nets refer to the electronic connection between cells.
We then obtain the geometric properties of each cell to form the physical layout of the entire circuit after placement, as shown in Figure 1 b.
Consider defining a single view graph data $\mathcal{G}=(\mathcal{V},\mathcal{E})$, where $v$ represents the sets of $\mathcal{N}$ nodes, $e_{ij} \in \mathcal{E}$ denotes the relationship between node $i$ and node $j$.
We thus construct our multi-view graph data with topological and geometrical properties.
From the topological view, the edge $(v, u)$ in the graph $\mathcal{G}_t$ represents the topology of cell connections stored in nets. Like~\cite{yangversatile}, the feature matrix $\mathcal{X}_t$ contains the handicraft features like pin density, node density, net density, etc.
From the geometrical view, the connection between cells depends on the layout structure. We use Delaunay triangulation algorithms~\cite{Delaunay} to build cell interactions in $\mathcal{G}_g$. Additional information such as coordinates and distance are used as auxiliary structure features in $\mathcal{X}_g$.

\subsection{HybridNet}
HybridNet, shown in Figure 1 c, can be described as having two networks with topological and geometrical pathways that focus on different views of graphs.

%%%
\noindent\textbf{Topology Pathway} can be any graph neural network that works on node message-passing.
Due to the input netlist can contain millions of nodes in VLSI circuit design, we hope to focus on the most relevant parts of the input to produce plausible representations.
The self-attention mechanism is the most promising method to approach this goal. Therefore, we stack $l$ layers of the Graph Attention Network~\cite{gat} as a topology aggregation module to instantiate this pathway.

%%%
\noindent\textbf{Geometric Pathway} consists of two functional modules, a position encoding module and a graph aggregation module.
Position encoding is the crucial part of the former module, which is a popular technique used in natural language processing by mapping spatial coordinates with sine and cosine functions~\cite{transformers}.
Instead of pre-computing the position encoding feature, we design a learnable position encoding module combined with a multi-layer perceptive network (MLP) to adaptively represent the geometric information and cell order.
In parallel to the topological pathway, we instantiate this pathway with continuous-filter convolutions which are the basic block of SchNet~\cite{schnet}.

%%%%
\noindent\textbf{Fusion Strategy}
After $l$ layers feature propagation, a simple yet effective fusion strategy is performed to aggregate multi-view graph features.
We concatenate the output of each pathway with its original features and run each through MLP.

Finally, the cell representation is obtained by concatenating the above features as input to the last MLP layer. 

%% file: doc/result.tex
\vspace{-.05in}
\section{Experiment}
\label{sec:Results}

To evaluate our methods, we perform experiments on 9 designs from the ISPD2015 competition benchmarks [13]. We use 6 for training and 3 for testing, where all designs in the training and testing sets have no overlap. 
To make the dataset more convincing, we use Cadence Innovus for placement and routing. The output congestion maps are treated as golden labels.
Following the protocol of \cite{yangversatile}, we use Pearson/Spearman/Kendall correlation metrics as the primary metrics to evaluate the the performance of the models.

%%%
In order to verify the generalisability of our method, we choose the following typical methods and their variant network as a basic baseline: GCN~\cite{gcn}, GAT~\cite{gat}, the two path variants of the above two methods, and the heterogeneous graph-based method NetlistGNN~\cite{yangversatile}.
We train the models with the AdamW optimizer 
for 500 epochs with an initial learning rate 2e-4.
The baseline methods GCN, GAT and their variants with the same number of layers start with the same graph data as our HybridNet. For NetlistGNN, we generate the heterogeneous graph with the same setting presented in \cite{yangversatile}. In addition, the experiments on pix2pix show that it cannot learn a generalised representation in such a small training set, so the results of vision-based methods are not presented in this paper. 
Table 1 shows the comparison with the baseline results for our HybridNet, the best and second-best scores of the baseline methods are \textbf{highlighted} and \underline{underlined}.
Obviously, our best model provides higher correlation accuracy compared to vanilla GCN, GAT and their variants. It shows that our HybridNet has a stronger ability to make accurate congestion predictions.
Furthermore, using the Pearson correlation metric, our model (52.2\%) is 10.9\% better than the best result presented in NetlistGNN (41.3\%), demonstrating that multi-view graphs have a better representational ability than both the single-view homograph and the heterogeneous graph.
\begin{table}[t]
\begin{center}

\caption{Experimental results on ISPD 2015 benchmarks \cite{nam2005ispd2005}.* denotes the architecture implemented by ourselves.}
\label{tab:ISPD2005Benchmark}
\vspace{-.05in}
{

\begin{tabular}{l|c|c|c}
\toprule[0.15em]
% \hline
    Method & Pearson & Spearman & Kendall \\  
    \midrule[0.15em]
    GCN~ & 0.290 & 0.250 & 0.203 \\  
    GAT~\cite{gat} & 0.131 & 0.005 & 0.004 \\  
    Two Pathways GCN* & 0.357 & 0.230 & 0.186 \\  
    Two Pathways GAT* & 0.308 & \textbf{0.273} & \textbf{0.221} \\  
    NetlistGNN*~\cite{yangversatile} & \underline{0.413} & 0.216 & 0.189 \\ 
    \textbf{HybridNet} (ours) & \textbf{0.522} & \underline{0.271} & \underline{0.220} \\ 
    \bottomrule[0.15em]
\end{tabular}

}
\vspace{-8mm}
\end{center}
\end{table}
%}}}

%% file: doc/conclu.tex
\vspace{-.05in}
\section{Conclusion}
\label{sec:Conclusion}

In this paper, we present a new perspective on the construction of a topological-geometric view graph by considering 
the nature of multi-views in circuit netlists.
We further propose HybridNet, a dual-branch network that fully aggregates two different types of graphs, accompanied by an effective fusion strategy to provide accurate congestion prediction.
The empirical study shows that our network can achieve significant improvements over conventional methods.
We hope that multi-view graph construction and aggregation network will foster further research in the EDA domain.

%% file: doc/acknowledgement.tex
\section*{acknowledgement}

This work was supported in part by the National Key Research and Development Program of China (No. 2021ZD0114702).

%% file: ISEDA2023_Routability.bbl
\begin{thebibliography}{10}
\providecommand{\url}[1]{#1}
\csname url@samestyle\endcsname
\providecommand{\newblock}{\relax}
\providecommand{\bibinfo}[2]{#2}
\providecommand{\BIBentrySTDinterwordspacing}{\spaceskip=0pt\relax}
\providecommand{\BIBentryALTinterwordstretchfactor}{4}
\providecommand{\BIBentryALTinterwordspacing}{\spaceskip=\fontdimen2\font plus
\BIBentryALTinterwordstretchfactor\fontdimen3\font minus
  \fontdimen4\font\relax}
\providecommand{\BIBforeignlanguage}[2]{{%
\expandafter\ifx\csname l@#1\endcsname\relax
\typeout{** WARNING: IEEEtran.bst: No hyphenation pattern has been}%
\typeout{** loaded for the language `#1'. Using the pattern for}%
\typeout{** the default language instead.}%
\else
\language=\csname l@#1\endcsname
\fi
#2}}
\providecommand{\BIBdecl}{\relax}
\BIBdecl

\bibitem{isola2017image}
P.~Isola, J.-Y. Zhu, T.~Zhou, and A.~A. Efros, ``Image-to-image translation
  with conditional adversarial networks,'' in \emph{Proc.~CVPR}, 2017, pp.
  1125--1134.

\bibitem{chai2022circuitnet}
Z.~Chai, Y.~Zhao, Y.~Lin, W.~Liu, R.~Wang, and R.~Huang, ``Circuitnet: An
  open-source dataset for machine learning applications in electronic design
  automation (eda),'' \emph{Science China Information Sciences}, vol.~65,
  no.~12, pp. 227\,401--, 2022.

\bibitem{routenet}
Z.~Xie, Y.-H. Huang, G.-Q. Fang, H.~Ren, S.-Y. Fang, Y.~Chen, and J.~Hu,
  ``Routenet: Routability prediction for mixed-size designs using convolutional
  neural network,'' in \emph{Proc.~ICCAD}.\hskip 1em plus 0.5em minus
  0.4em\relax IEEE, 2018, pp. 1--8.

\bibitem{liu2021global}
S.~Liu, Q.~Sun, P.~Liao, Y.~Lin, and B.~Yu, ``Global placement with deep
  learning-enabled explicit routability optimization,'' in
  \emph{Proc.~DATE}.\hskip 1em plus 0.5em minus 0.4em\relax IEEE, 2021, pp.
  1821--1824.

\bibitem{high-definition}
M.~B. Alawieh, W.~Li, Y.~Lin, L.~Singhal, M.~A. Iyer, and D.~Z. Pan,
  ``High-definition routing congestion prediction for large-scale fpgas,'' in
  \emph{Proc.~ASPDAC}.\hskip 1em plus 0.5em minus 0.4em\relax IEEE, 2020, pp.
  26--31.

\bibitem{congestionnet}
R.~Kirby, S.~Godil, R.~Roy, and B.~Catanzaro, ``Congestionnet: Routing
  congestion prediction using deep graph neural networks,'' in
  \emph{Proc.~VLSI-SoC}.\hskip 1em plus 0.5em minus 0.4em\relax IEEE, 2019, pp.
  217--222.

\bibitem{congestionnetnn}
C.~Ma, Y.~Xiao, S.~Wang, J.~Yu, and J.~Chen, ``Congestnn: An bi-directional
  congestion prediction framework for large-scale heterogeneous fpgas,'' in
  \emph{Proc.~ASICON}.\hskip 1em plus 0.5em minus 0.4em\relax IEEE, 2021, pp.
  1--4.

\bibitem{yangversatile}
S.~Yang, Z.~Yang, D.~Li, Y.~Zhang, Z.~Zhang, G.~Song, and H.~Jianye,
  ``Versatile multi-stage graph neural network for circuit representation,'' in
  \emph{Proc.~NeurIPS}, 2022.

\bibitem{survey}
C.~Xu, D.~Tao, and C.~Xu, ``A survey on multi-view learning,'' \emph{arXiv
  preprint arXiv:1304.5634}, 2013.

\bibitem{Delaunay}
D.-T. Lee and B.~J. Schachter, ``Two algorithms for constructing a delaunay
  triangulation,'' \emph{International Journal of Computer \& Information
  Sciences}, vol.~9, no.~3, pp. 219--242, 1980.

\bibitem{gat}
P.~Veli{\v{c}}kovi{\'c}, G.~Cucurull, A.~Casanova, A.~Romero, P.~Lio, and
  Y.~Bengio, ``Graph attention networks,'' in \emph{Proc.~ICLR}, 2017.

\bibitem{transformers}
D.~Lu, Q.~Xie, M.~Wei, L.~Xu, and J.~Li, ``Transformers in 3d point clouds: A
  survey,'' \emph{arXiv preprint arXiv:2205.07417}, 2022.

\bibitem{schnet}
K.~Sch{\"u}tt, P.-J. Kindermans, H.~E. Sauceda~Felix, S.~Chmiela,
  A.~Tkatchenko, and K.-R. M{\"u}ller, ``Schnet: A continuous-filter
  convolutional neural network for modeling quantum interactions,'' in
  \emph{Proc.~NeurIPS}, 2017.

\bibitem{gcn}
T.~N. Kipf and M.~Welling, ``Semi-supervised classification with graph
  convolutional networks,'' \emph{Proc.~ICLR}, 2016.

\bibitem{nam2005ispd2005}
G.-J. Nam, C.~J. Alpert, P.~Villarrubia, B.~Winter, and M.~Yildiz, ``The
  ispd2005 placement contest and benchmark suite,'' in \emph{Proceedings of the
  2005 international symposium on Physical design}, 2005, pp. 216--220.

\end{thebibliography}
